\documentclass[
]{ceurart}

\usepackage{hyperref}
\usepackage{listings}
\makeatletter
\newcommand{\rmnum}[1]{\romannumeral #1}
\newcommand{\Rmnum}[1]{\expandafter\@slowromancap\romannumeral #1@}
\makeatother
\begin{document}

\copyrightyear{2024}
\copyrightclause{Copyright for this paper by its authors. Use permitted under Creative Commons License Attribution 4.0 International (CC BY 4.0).}

\conference{EAmSI24: Edge AI meets swarm intelligence, September 18, 2024, Dubrovnik, Croatia}

\title{Co-Learning: Towards Semi-Supervised Object Detection with Road-side Cameras}


\author[1]{Jicheng Yuan}[%
orcid=0009-0002-4448-2809,
email=jicheng.yuan@tu-berlin.de
]

\address[1]{Open Distributed Systems (ODS), Technical University of Berlin, Berlin 10623, Germany}

\author[1]{Anh Le-Tuan}[%
orcid=0000-0003-2458-607X,
email=anh.letuan@tu-berlin.de,
]

\author[1]{Ali Ganbarov}[%
orcid=0009-0000-7543-2391,
email=ali.ganbarov@tu-berlin.de,
]
\author[1,2]{Manfred Hauswirth}[%
orcid=0000-0002-1839-0372,
email=manfred.hauswirth@tu-berlin.de,
]
\author[1,2]{Danh Le-Phuoc}[%
orcid=0000-0003-2480-9261,
email=danh.lephuoc@tu-berlin.de,
]
\address[2]{Fraunhofer-Institut für Offene Kommunikationssysteme (FOKUS), Berlin 10589, Germany}


\begin{abstract}
Recently, deep learning has experienced rapid expansion, contributing significantly to the progress of supervised learning methodologies. However, acquiring labeled data in real-world settings can be costly, labor-intensive, and sometimes scarce. This challenge inhibits the extensive use of neural networks for practical tasks due to the impractical nature of labeling vast datasets for every individual application. To tackle this, semi-supervised learning (SSL) offers a promising solution by using both labeled and unlabeled data to train object detectors, potentially enhancing detection efficacy and reducing annotation costs. Nevertheless, SSL faces several challenges, including pseudo-target inconsistencies, disharmony between classification and regression tasks, and efficient use of abundant unlabeled data, especially on edge devices, such as roadside cameras. Thus, we developed a teacher-student-based SSL framework, Co-Learning, which employs mutual learning and annotation-alignment strategies to adeptly navigate these complexities and achieves comparable performance as fully-supervised solutions using 10\% labeled data.
\end{abstract}


\maketitle

\section{Introduction}
Object detection has significantly advanced with the advent of deep learning techniques~\cite{zou2023object}. While current approaches perform exceptionally well with ample labeled data and computing resources, they often falter in data-limited scenarios, especially on edge devices. Costly, labor-intensive dataset preparation poses significant challenges for autonomous driving models, particularly in urban areas or where roadside cameras are dense. Previous studies~\cite{qu2020privacy, jiang2021flexible} primarily focused on supervised learning with synthetic data or on edge devices, both requiring labeled data from target scenarios. However, obtaining labeled data for every possible use case is resource-intensive and often impractical, leading to the emergence of semi-supervised learning (SSL) as a potential solution.

\begin{figure*}[htbp]
    \centering
    \includegraphics[width=1.0\textwidth]{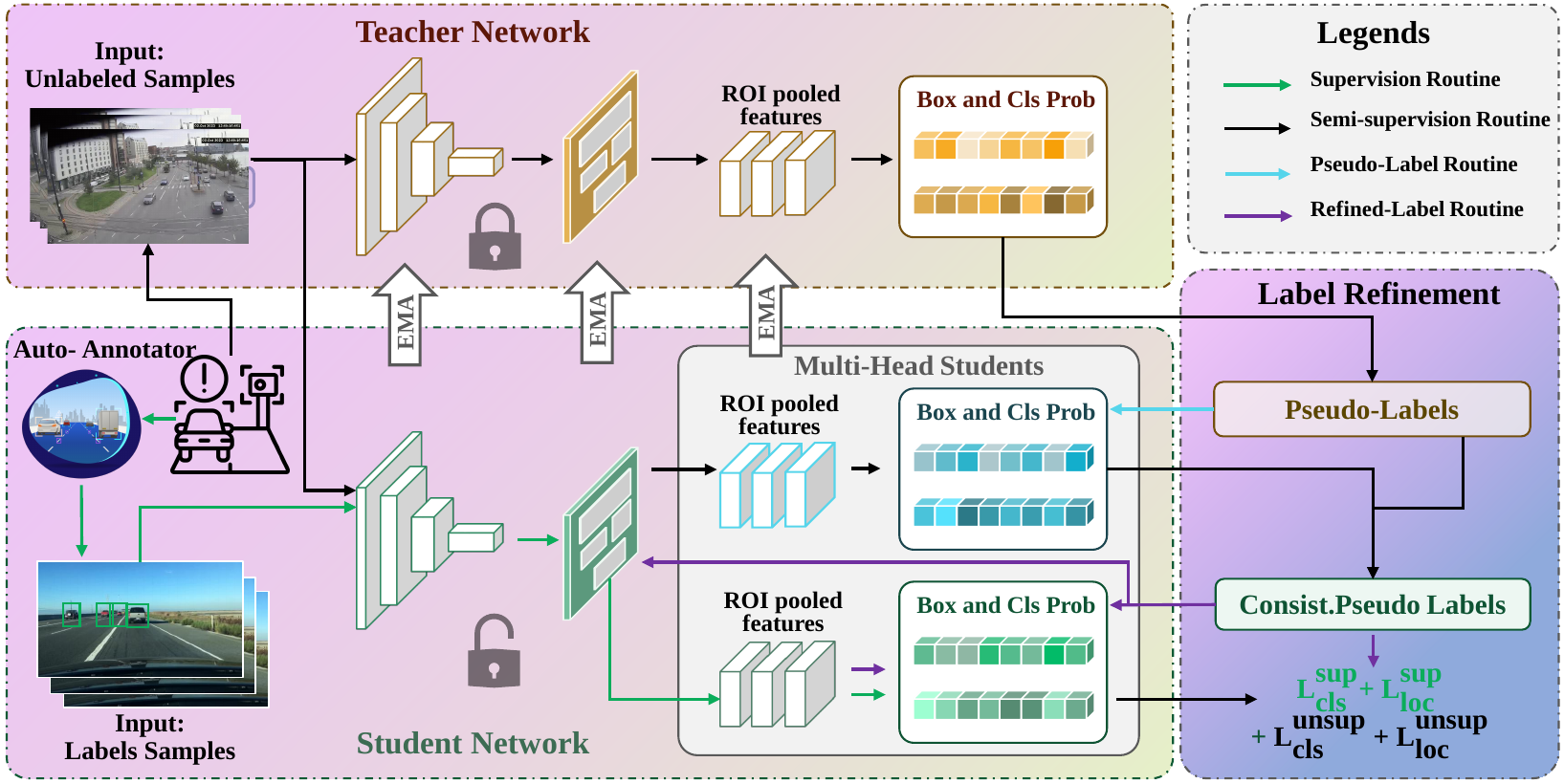}
    \caption{Overview of the proposed Co-Learning framework: three modules to address the inconsistency in SSOD on computing-power-constrained devices, $\left(\rmnum{1}\right)$ Pseudo-labels are determined by dynamic thresholds; $\left(\rmnum{2}\right)$ Consistent pseudo-labels across teacher and student networks contribute to aligning and calibrating regression quality; $\left(\rmnum{3}\right)$ Multi-head student networks reassign anchors based on matching cost.}
    \label{fig:consistent_teacher}
    \vspace{-5mm}
\end{figure*}

Hence, to boost the application of Semi-supervised Object Detection (SSOD) methods on edge devices, we propose a novel learning pipeline Co-learning, from data curation to learning execution, as delineated in Figure~\ref{fig:consistent_teacher}, to mitigate the challenges inherent in SSOD, such as data inconsistency. Adhering to the constraints set by the AI CITY CHALLENGE dataset~\cite{Naphade23AIC23}, we employ data from Track 2. Our goal is to bridge the existing gap in limited data and resources by introducing a semi-supervised learning strategy for object detection. This strategy intends to enhance the utilization of unlabeled data during the learning phase, leveraging both labeled and unlabeled data to improve object detection with generalized features.

To achieve this, based on a teacher-student network (TSN), we start with 10\% of the data from AI City Challenge~\cite{Naphade23AIC23} Track 2 for supervised learning. The remaining 90\% of the data is used as unlabeled data to train the model in a semi-supervised manner. However, in unsupervised TSN, a small deviation in the teacher network will result in significant noise at the boundaries of pseudo-boxes, causing erroneous targets to be associated with nearby objects under static IoU-based assignment. This owing to some inactivated anchors being falsely assigned as positive in the student network and will guide the overfitting of the student network, as well as producing inconsistent labels for neighboring objects, that belong to the same category. By employing refined, consistent pseudo-labels, we ensure that the student network learns from a set of pseudo-labels that are more aligned with true situations, enhancing overall object detection performance.

The remainder of the paper is structured as follows: Section~\ref{sec:rw} discusses related work. Section~\ref{sec:pipeline} details the data cleaning process and proposed learning pipeline, incorporating stemming and lemmatization for text descriptions and the SSOD approach. Section~\ref{sec:exper} outlines the experimental setup and presents initial results of the proposed learning framework. Finally, Section~\ref{sec:con} concludes the paper and presents the next step of the proposed framework.

\section{Related Work} \label{sec:rw}

\subsection{Object Detection}
Object detection has been substantially influenced by the rapid advancement of deep learning techniques, which is a task that not only identifies the class of objects but also localizes them within images or videos~\cite{zou2023object}. Its pipeline is similar to image classification but includes some additional steps, such as data acquisition with corresponding bounding boxes around the objects of interest and labels indicating the class of each object. Modern object detectors mainly consist of two principal architectures: single-stage and two-stage models. Two-stage detector models, such as Faster-RCNN (FRCNN)~\cite{ren2015faster} , are characterized by an integrated framework that includes a region proposal network (RPN), alongside a classification and regression head. In this way, the backbone extracts features from the input image, while the region proposal algorithm generates potential object bounding boxes. These features are then passed through the detection heads to perform object classification and localization. On the other hand, one-stage models, such as YOLO~\cite{redmon2016you}, SSD~\cite{liu2016ssd}, and RetinaNet~\cite{lin2017focal}, generate bounding box predictions and class probabilities without an RPN. Additionally, transformer-based models have been applied to object detection. The DEtection TRansformer (DETR)~\cite{carion2020end} uses a transformer to process the input image and generate object proposals, which are then refined using a feedforward network. Object detection has many applications, such as identifying pedestrians in self-driving cars~\cite{zou2023object}, detecting objects in surveillance footage~\cite{loganathan2019suspicious}, and identifying defective objects in automatic production assembly lines~\cite{benbarrad2021intelligent}. In this work, we employ Faster-RCNN~\cite{ren2015faster} as the base detector for both the teacher and student networks.

\subsection{Semi-supverised Object Detection}
In SSOD, a prevalent approach involves the generating of pseudo-bounding boxes through a teacher model, with the anticipation that student detectors will yield uniform predictions on enhanced input samples~\cite{jeong2019consistency, li2020improving, liu2021unbiased}. Traditionally, two-stage detectors~\cite{jeong2019consistency,liu2021unbiased, xu2021end} have led the way in conventional SSOD techniques, showcasing their dominance. However, single-stage detectors~\cite{chen2022dense,zhang2022s4od,zhou2022dense} have also emerged as formidable contenders, noted for their straightforward design and superior performance. In this work, considering the localization quality of potential objects, we utilize a two-stage teacher-student SSOD approach, primarily addressing the issues of inconsistency present therein, which involves adaptive proposal assignment, pseudo label refinement and alignment, and a dynamic threshold mechanism, all aimed at enhancing label quality with less annotation cost.

\section{Methodology and Experiments} \label{sec:pipeline}
In this section, we will present the details of the method to execute our proposed Co-Learning pipelines on the 7th AI-City-Challenge Dataset. It comprises three main steps, each addressing the task’s challenges with varying levels of detail, and achieving remarkable performance on the benchmark dataset. The three steps are $\left(\rmnum{1}\right)$ data curation and annotation alignment, $\left(\rmnum{2}\right)$ proposed learning pipeline and configuration, and $\left(\rmnum{3}\right)$ overall training platform.

\subsection{Data Curation and Annotation Alignment} \label{sec:alignment}
To enhance the development of text-to-image retrieval systems, particularly in object detection, the 7th AI City Challenge~\cite{Naphade23AIC23} track two introduced a dataset for tracked vehicle retrieval using natural language descriptions, offering diverse textual descriptions. However, these annotations lack box-level descriptions for specific objects. Previous studies~\cite{xie2023unified, le2023tracked} have indeed shown encouraging outcomes, but numerous challenges persist. One primary concern is the inherent complexity of natural textual data. While humans can effortlessly comprehend textual narratives, machines struggle to differentiate between similar descriptions, such as \textit{A van is moving straight} and \textit{The red van is heading forward}. The scarcity of training data further intensifies this challenge for machine learning models. Another significant obstacle is the lack of abundant high-quality training data tailored for text-to-image vehicle retrieval, given the early stage of this research. Compared to well-established datasets, such as Waymo Open Dataset~\cite{sun2020scalability} and COCO~\cite{lin2014microsoft}, which boast millions of samples, learning on edge devices suffers from limited high-quality annotations. Given the complexity of real-world road scenes, especially from fixed camera perspectives, it is essential to extract diverse features from unlabeled data for further refinement. Therefore, we selected the AI City Challenge dataset as our initial point. To assign each box a specific noun description and extract consistent box-level annotations, we applied stemming and lemmatization to text descriptions, inspired by the \cite{le2023tracked} approach. These are standard techniques in Natural Language Processing (NLP). While stemming truncates words to their root forms, lemmatization derives a word's base form considering language grammar. These techniques aid in stop word removal, correction of misspelled words, and conversion to consistent base forms. Addressing linguistic ambiguities in queries and ensuring uniform textual embeddings is essential. The vehicle descriptions typically encompass three primary attributes: color, type, and movement. By employing the English PropBank Semantic Role Labeling (SRL) method~\cite{palmer2011semantic}, we extracted these attributes and proposed a standardized format $text_{standardized}=attr_c+attr_t+attr_m$, where $attr_c$, $attr_t$, and $attr_m$ represent vehicle color, type, and motion, respectively. This identified synonymous terms with equivalent semantic meanings. To ensure consistent box-level annotation and minimize the complexity of the learning aim, we then grouped these synonyms into clusters based on semantic similarity and replaced them with a representative term. For instance, terms such as \textit{red van}, \textit{van}, and \textit{blue van} are grouped under the label \textit{van}.

\begin{figure*}[htbp]
    \centering
    \includegraphics[width=1.0\textwidth]{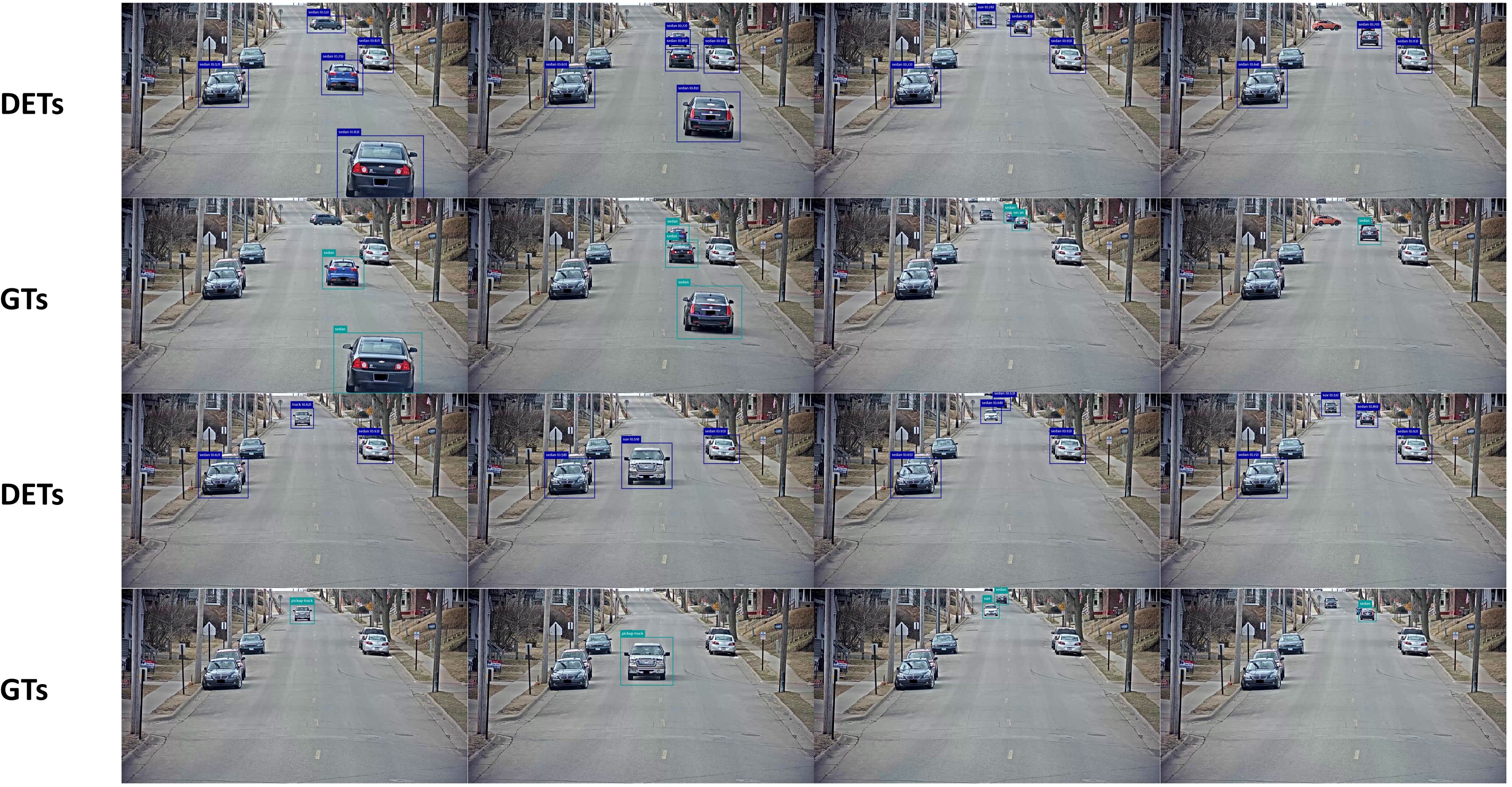}
    \caption{Qualititive Analysis of the proposed Co-Learning solution on the 7th AI CITY CHALLENGE Track 2 dataset. The base detector is Faster-RCNN with ResNet-50 as the feature extractor for all experiments to ensure a fair comparison. (\textbf{DETs}: \textbf{Detected Results}, \textbf{GTs}: \textbf{Ground Truths})}
    \label{fig:results_det}
    \vspace{1mm}
\end{figure*}

\subsection{Overall Training Platform}
All experiments were conducted on a dedicated server equipped with two Nvidia Tesla V100‐16GB GPUs, ensuring optimal performance and parallel processing capabilities. The server features 4 $\times$ Intel(R) Xeon(R) Gold 5117 CPU, complemented by 2TB of DDR4 RAM. The training time for Co-Learning, combined with the initial hyperparameter search conducted using Bayesian Optimization~\cite{snoek2012practical}, amounted to approximately 90 GPU hours. In addition, we employed PyTorch 1.9.0 and MMDetection 2.25.0 as the primary framework and toolbox.

\begin{table*}[ht]
\centering
\scalebox{0.88}{
\begin{tabular}{ccccccccc}
\toprule[0.8pt]
\textbf{Methods} & \textbf{Sedan} & \textbf{Bus} & \textbf{Pickup-Truck} & \textbf{SUV} & \textbf{Hatchback} & \textbf{Van} & \textbf{Truck} & \textbf{mAP} \\ \hline
Oracle       &     26.2          & 76.6            & 31.8                     & 17.6            & 0.6                  & 29.5            & 0.5              & 36.1                \\ \hline
w/o Annotation-Alignment  & 14.4           & 73.0         & 14.2                  & 10.0         & 0.0                & 5.4          & 0.0            & 23.0             \\
w/ Annotation-Alignment    & 23.8           & 75.2         & 36.1                  & 22.3         & 0.1                & 31.8         & 1.1            & 36.5             \\ \hline
Improvement vs. Oracle    & \textcolor{blue}{-2.4\%}           & \textcolor{blue}{-1.4\%}          & \textcolor{blue}{+4.3\%}                   & \textcolor{blue}{+4.7\%}          & \textcolor{blue}{-0.5\%}                & \textcolor{blue}{+2.3\%}         & \textcolor{blue}{+0.6\%}            & \textcolor{blue}{+0.4\%}             \\
\bottomrule[0.8pt]
\end{tabular}}
\caption{Quantitative analysis across the proposed Co-Learning solution on AI CITY CHALLENGE dataset, strategies with (w/) and without (w/o) annotation alignment, as well as the oracle model (which can access all of the pseudo-labels). We reported $AP^{0.5}$ as the mean averaged precision $mAP$.}
\label{training_results}
\end{table*}

\subsection{Experimental setup and Results}
Given the identified challenges from abundant unlabeled data and complex road scene scenarios acquired by roadside cameras, we utilize a semi-supervised object detection solution in a teacher-student network. This methodical strategy aims to reduce pseudo-label inconsistencies generated by the teacher model. Our approach incorporates three primary modules to counteract this. Initially, the dynamic pseudo-label assignment module replaces the traditional IoU-based strategy, enhancing the student network’s resilience against noisy pseudo-bounding boxes. Next, the pseudo-feature alignment module calibrates subtask predictions, allowing each classification feature to adaptively choose the most suitable feature vector for the regression task, regardless of its scale or location. Then, the pseudo-label refinement module dynamically adjusts the score threshold for pseudo-bounding boxes, ensuring stability in the count of ground truths during the early training stages and rectifying any unreliable supervision signals. Initial results of our proposed solution on roadside scenarios, mainly on the 7th AI City Challenge Track 2 dataset), have demonstrated its effectiveness as shown in Table~\ref{training_results}. Notably, using only 10\% of annotated data with a ResNet-50 backbone, without annotation alignment, it achieves a mean average precision (mAP) of 23.0. When further trained on the fully annotated dataset with additional unlabeled data and the annotation alignment strategy, the performance escalates to an mAP of 36.5, surpassing the oracle model trained only on pseudo labels with an improvement of 0.4\% on overall mAP.

\subsection{Further Analysis} \label{sec:exper}
In all our experiments, we utilized the Co-Learning method, depicted in Figure~\ref{fig:consistent_teacher}, as our SSOD design, implemented using PyTorch. To ensure better clarity and consistency, we adopted the official FRCNN~\cite{ren2015faster} model from TorchVision~\cite{torchvision2016} and fixed the learned parameters in the ResNet-50~\cite{he2016deep} backbone.
In terms of dataset configuration and adhering to the framework's guidelines, we used a total of 10\% of the annotated data for the initial supervised training, treating the remaining 90\% as unlabeled. We utilized the mean average precision (mAP) across object classes for each video frame as the evaluation metric. For a comprehensive assessment, we also provided the class-wise AP.
For training, we utilized the standard implementation, configuring the Co-Learning to adapt to our training scenarios. 
We initialized the learning rate, momentum, and weight\_decay at $1e-2$, $9e-1$, and $1e-4$, respectively, and conducted training using a batch size of five on a single GPU, performing $180K$ iterations.
In terms of method comparison, our approach benchmarked two methods: with and without annotation alignment, emphasizing that reducing the complexity of annotations could decrease model performance. Further details are provided in Section~\ref{sec:alignment}.
As indicated in Table~\ref{training_results}, the proposed annotation alignment strategy outperformed its counterpart Oracle model with an improvement of 0.4\% mAP. We hypothesize that issues with label consistency could limit performance, which could be solved by employing stemming, lemmatization, and label-consistent strategies. Figure~\ref{fig:results_det} showcases some examples of the final detected objects, which indicates that the proposed architecture can generalize well on target features, even if labeled data is limited.

\section{Conclusions} \label{sec:con}
In this work, we investigate how to achieve consistent labeling using the proposed Co-learning framework and transfer it to the datasets for which annotations are limited. We employ text stemming and lemmatization methods to decrease the complexity of annotations towards semi-supervised object detection in road scenarios. Relative to baseline methods without annotation alignment, our findings suggest the importance of label consistency in SSOD. Next, we will transfer the proposed architecture and learned knowledge to edge devices, such as Jetson Orin and Xavier. In addition, benefiting from the advance of visual language models and visual agents, leaning on the edge efficiently would further widen the capability of the proposed Co-Learning framework. 

\section{Acknowledgements}
This work is supported by the German Research Foundation (DFG) under the COSMO project (grant No. 453130567), and by the European Union’s Horizon WIDERA under the grant agreement No. 101079214 (AIoTwin),  by the Federal Ministry for Education and Research Germany under grant number 01IS18037A (BIFOLD), and RIA research and innovation program under the grant agreement No. 101092908 (SmartEdge).

\bibliography{sample-ceur}

\end{document}